\documentclass[letterpaper,journal]{IEEEtran}

\IEEEoverridecommandlockouts                              

\usepackage{amsmath,amsfonts}
\usepackage{algorithmic}
\usepackage{algorithm}
\usepackage{array}
\usepackage{bbding}
\usepackage{color}
\usepackage{booktabs}
\usepackage{threeparttable}
\usepackage{textcomp}
\usepackage{stfloats}
\usepackage{url}
\usepackage{verbatim}
\usepackage{graphicx}
\usepackage{cite}
\usepackage[caption=false,font=normalsize,labelfont=sf,textfont=sf]{subfig}

\makeatletter
\let\NAT@parse\undefined
\makeatother
\usepackage{hyperref}  

\begin{document}

\title{\LARGE \bf MAD: \mbox{Mapping-Aware} World Models for\\
Agile Quadrotor Flight}

\author{
Xinhong Zhang, 
Runqing Wang, 
Yunfan Ren,
Ding Yu,
Boyu Zhou,
Jian Sun,
\\
Fang Deng,~\IEEEmembership{Fellow,~IEEE},
Jie Chen,~\IEEEmembership{Fellow,~IEEE},
and Gang Wang,~\IEEEmembership{Senior Member, IEEE} 
\thanks{This work was supported in part by the National Natural Science Foundation of China under Grants U23B2059, 62088101, and also by the Zhongguancun Academy under Grant 20240307. (\emph{Xinhong Zhang and Runqing Wang contributed equally to this paper; Corresponding author: Gang Wang.})}
\thanks{Xinhong Zhang is with the State Key Lab of Autonomous Intelligent Unmanned Systems, Beijing Institute of Technology, Beijing 100081, China, and also with the Zhongguancun Academy, Beijing 100094, China. E-mail: xhzhang@bit.edu.cn.}
\thanks{Runqing Wang, Ding Yu, Jian Sun, Fang Deng, and Gang Wang are with the State Key Lab of Autonomous Intelligent Unmanned Systems, Beijing Institute of Technology, Beijing 100081, China. E-mail: bitwrq@bit.edu.cn, yd@bit.edu.cn, sunjian@bit.edu.cn, 
dengfang@bit.edu.cn, 
gangwang@bit.edu.cn.}
\thanks{Yunfan Ren is with the School of Computer Science and Technology, Tongji University. E-mail: yunfan@tongji.edu.cn.}
\thanks{Boyu Zhou is with the Department of Mechanical and Energy Engineering, Southern University of Science and Technology, Shenzhen 518055, China. E-mail: zhouby@sustech.edu.cn.} 
\thanks{Jie Chen is with the Harbin Institute of Technology, and also with the State Key Lab of Autonomous Intelligent Unmanned Systems, Beijing Institute of Technology, Beijing 100081, China. E-mail: chenjie@bit.edu.cn.}
}


\maketitle
\thispagestyle{empty}
\pagestyle{empty}

\allowdisplaybreaks

\begin{abstract}
Agile quadrotor flight in cluttered scenes requires more than a reactive mapping from a depth image to a control command: the vehicle must remember which regions have been observed, infer nearby occupied space, and act under partial visibility and tight latency. Classical aerial navigation stacks provide this structure with explicit odometry, mapping, planning, and tracking modules, but the interfaces between these modules can introduce latency, error accumulation, and substantial engineering overhead. End-to-end learning-based policies remove these interfaces, yet they often lack interpretable spatial memory and transfer poorly across environments and tasks. In this paper, we present \emph{Mapping-Aware Dreamer (MAD)}, a geometry-aware world model for vision-based quadrotor flight. Instead of using raw-image reconstruction as the main self-supervised objective, MAD learns recurrent latent dynamics that reconstruct robocentric occupancy and visibility grid maps together with proprioceptive states. This design forces the latent state to encode local geometry, visibility history, and ego-motion in a form that is directly relevant to collision avoidance. MAD is trained in DiffAero using a GPU-parallel map-construction module that provides high-throughput supervision for occupancy and visibility. The learned representation is used in three policy-learning modes: imagination-based MAD-Dreamer and feature-extractor variants based on PPO and SHAC. Across visual navigation and racing tasks, MAD-based agents achieve higher success rates, faster flight, and better cross-task transfer than corresponding vision-only baselines. The model also produces interpretable map predictions and accurate ego-motion estimates from depth observations. We further deploy the learned policy on a physical quadrotor with an Intel RealSense D435i and demonstrate safe indoor and outdoor flight under limited sensing, reaching 9.66 m/s in simulation and 5.05 m/s in real-world forest experiments. These results show that mapping-aware world models provide a practical middle ground between modular aerial navigation and end-to-end learning.
\end{abstract}

\section{Introduction}

Agile quadrotor flight in cluttered environments is a tightly coupled perception, memory, and control problem. A vehicle must infer free and occupied space from a narrow onboard field of view, remember regions that were observed in the recent past, and produce dynamically feasible commands under strict sensing and compute latency. Classical unmanned aerial vehicle (UAV) stacks address this problem through visual--inertial or LiDAR odometry, local mapping, trajectory planning, and tracking control \cite{diffflatvijay, super}. This modular structure is interpretable and has enabled impressive real-world flight, but the interfaces between modules can introduce latency, accumulate errors, and require substantial hand-designed engineering.

\begin{figure}[!t]
\centering
\includegraphics[width=0.48\textwidth]{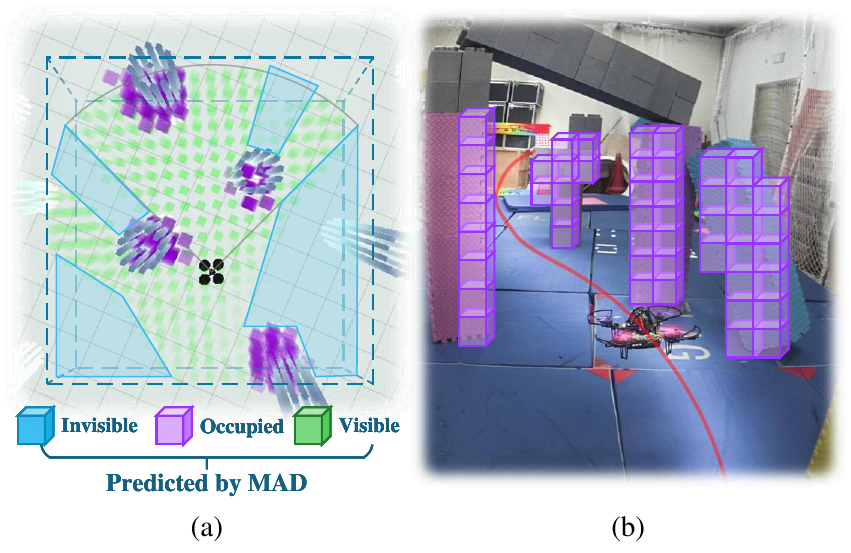}
\caption{(a) Visualization of the flight scene together with the occupancy and visibility grid maps predicted by MAD. Occupied voxels are shown in purple, while visible free-space voxels are shown in green. (b) Fast autonomous visual navigation in a cluttered indoor environment using an Intel RealSense D435i. During flight, MAD continuously predicts the occupancy and visibility of surrounding voxels.}\label{head}
\end{figure}

RL-based agile flight has recently achieved fast visual navigation, differentiable-physics training, and near-limit drone racing by optimizing perception and control jointly \cite{navrl, Zhang2025, yopo2025, championlevel, rlvsoptim}. These methods can produce fast flight without an explicit mapping-and-planning pipeline. However, a purely monolithic policy has no built-in notion of what space is occupied, visible, or unknown. This makes it difficult to diagnose failures, reuse the representation across tasks, or reason about safety when the depth image is partial and temporally aliased.

We therefore study an intermediate design between classical modular autonomy and fully end-to-end learning. General world models, including the Dreamer series and scalable latent-dynamics methods \cite{dreamerv1,dreamerv2,hafner2025dreamerv3,tdmpc2,wu2022daydreamer}, have shown that compact predictive states and imagined rollouts can improve RL from high-dimensional observations. For quadrotor flight, however, the learned state must be spatially grounded: it should predict not only future observations or rewards, but also the local geometry that determines collision risk. In contrast to prior learned modules \cite{zhang2025corb, AgilityAdaptation}, we use a latent world model as the interface between perception and control and supervise it directly with occupancy and visibility.

In this work, we propose \textbf{Mapping-Aware Dreamer} (\textbf{MAD}), a geometry-aware world model for vision-based quadrotor flight. MAD reconstructs robocentric occupancy grid maps (OGMs), visibility grid maps (VGMs), and proprioceptive states from onboard depth and sensory measurements. Compared with raw depth reconstruction, OGM/VGM supervision separates obstacle geometry from camera visibility and encourages the recurrent latent state to encode spatial memory and ego-motion. To train this model efficiently, we extend the simulator DiffAero \cite{zhang2025diffaero} with a fully GPU-parallel occupancy and visibility map-construction module.

The learned MAD representation can be leveraged by different policy learning algorithms. We instantiate an imagination based controller which we term MAD-Dreamer, and two feature-extractor variants, MAD-PPO and MAD-SHAC, based on proximal policy optimization (PPO) \cite{schulman2017proximal} and short-horizon actor-critic (SHAC) \cite{xu2021accelerated}. The experiments are organized accordingly: reconstruction evaluates the latent geometry, training and racing evaluate policy utility, and Gazebo/PX4 plus real flights evaluate onboard safety and agility. The resulting system is exported as a single onboard policy that maps depth images and visual--inertial odometry estimates to acceleration commands for real-time flight.

In a nutshell, the main contributions of this work are summarized as follows.
\begin{enumerate}
\item We propose MAD, a mapping-aware world model that reconstructs local occupancy, visibility, and proprioceptive states from visual observations, yielding interpretable and temporally consistent latent representations for agile quadrotor navigation.
\item We implement a GPU-parallel OGM/VGM construction module in DiffAero, achieving occupancy and visibility evaluation of $4.84 \times 10^8$ voxels per second on a single GPU and enabling high-throughput world model  training.
\item We validate MAD with Dreamer-style imagination learning, PPO, and SHAC, and demonstrate in simulation and real flight that MAD-based policies improve visual navigation and task transfer while reaching 9.66 m/s in simulation and 5.05 m/s in real-world forest experiments.
\end{enumerate}

\section{Related Work}

\subsection{Learning-based Autonomous Flight}
Learning-based methods have substantially improved the autonomy and agility of aerial robots. Imitation learning provides a direct way to map partial perceptual observations to expert actions and has been applied to navigation \cite{imitationnav}, racing \cite{imitationracing1,imitationracing2}, and swarm flight \cite{imitationswarm}. Its main limitation is that performance depends strongly on the coverage and quality of expert demonstrations, which is difficult to guarantee for aggressive flight in cluttered environments.

Model-free RL avoids expert demonstrations by optimizing control policies through interaction, and has shown strong performance in visual navigation \cite{depthrl1,depthrl2,rgbrl1} and racing \cite{championlevel,rlvsoptim,racing-survey}. However, purely model-free policies typically require extensive data and provide limited insight into the spatial information used for collision avoidance.

Differentiable simulation offers another route by propagating gradients through physical dynamics instead of treating the environment as a black box. Such methods can improve sample efficiency, sim-to-real transfer, and physical consistency \cite{Zhang2025,opticalflow,zhang2025diffaero}. Nevertheless, when the policy directly maps high-dimensional observations to commands, the resulting representation remains difficult to interpret.

world model based approaches seek to improve both efficiency and interpretability by learning predictive latent dynamics for physical states and high-dimensional observations. Dream to Fly \cite{dream2fly} learns a visual world model from RGB observations for agile maneuvering, while SkyDreamer \cite{skydreamer} uses privileged targets to make the learned representation easier to inspect. Existing aerial world models, however, still mainly emphasize temporal prediction or task reward. They do not explicitly impose the local geometric structure--occupied, visible, and unknown space--that is central to safe quadrotor navigation in cluttered environments.

\subsection{Model-based Reinforcement Learning}
Early model-based reinforcement learning (MBRL) learned explicit dynamics models for planning and data-efficient control, but these models often suffered from bias and compounding errors during long-horizon rollout. Latent world models have greatly improved the scalability of MBRL. The Dreamer family \cite{dreamerv1,hafner2025dreamerv3} learns recurrent latent dynamics from high-dimensional observations and optimizes policies on imagined trajectories. STORM \cite{storm} uses transformers to capture long-range dependencies, while MuZero \cite{muzero}, EfficientZero \cite{efficientzero}, TD-MPC \cite{tdmpc}, and TD-MPC2 \cite{tdmpc2} combine learned dynamics with planning or value prediction. Diffusion-based world models \cite{diamond} further emphasize high-fidelity visual rollout. These advances suggest that compact latent dynamics can support sample-efficient decision making. For real quadrotor flight, however, a useful world model must also be spatially grounded, robust to partial observations, and compatible with onboard sensing and computation \cite{wu2022daydreamer}.

\begin{figure*}[!t]
\centering
\includegraphics[width=\textwidth]{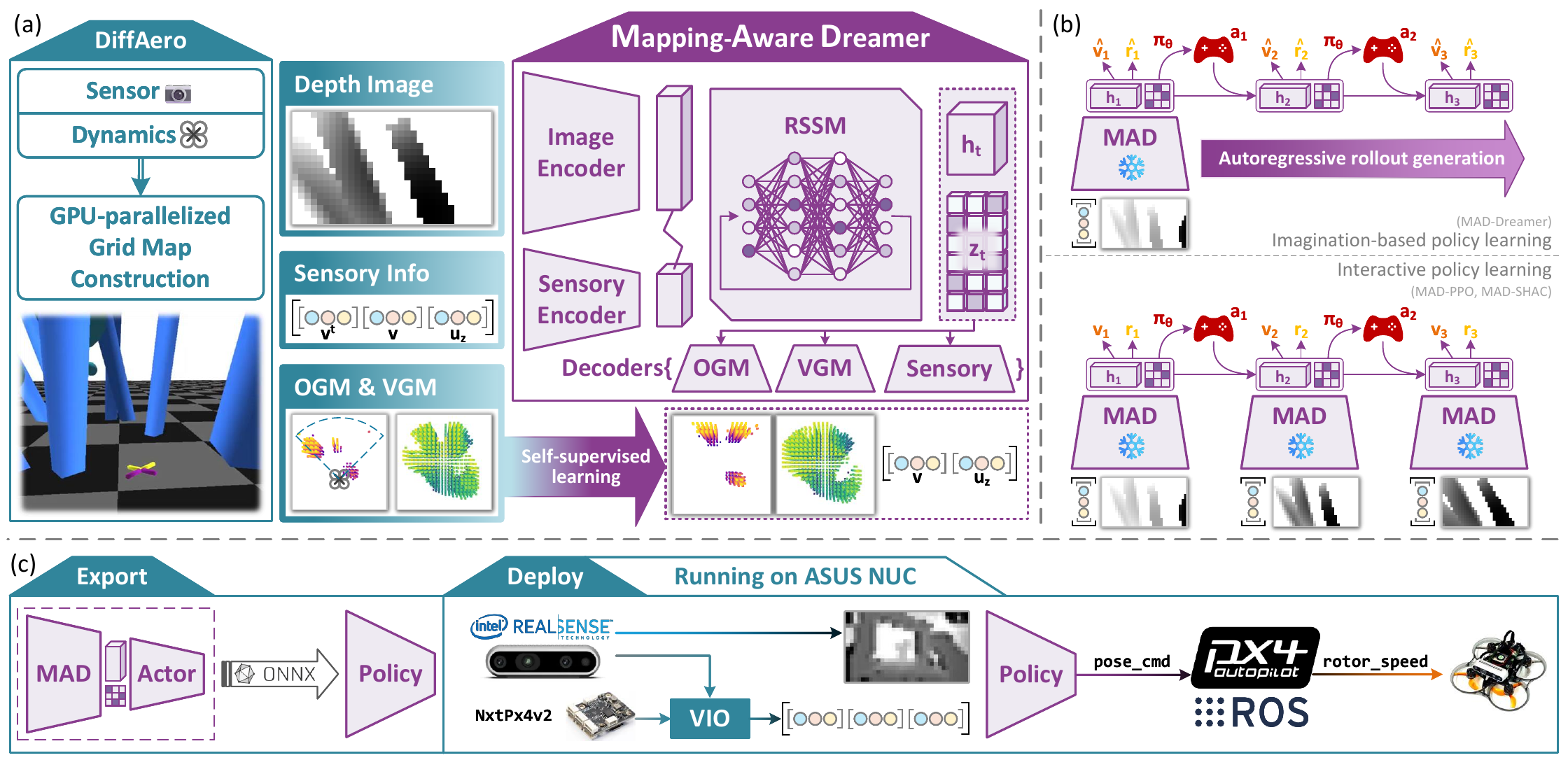}
\caption{\textbf{Overview of Mapping-Aware Dreamer (MAD) and the proposed MAD-based visual navigation system.} (a) Self-supervised training of MAD in the GPU-based DiffAero simulator. DiffAero provides depth images, proprioceptive sensory information, and robocentric occupancy and visibility grid maps, and MAD is trained to infer latent states $h_t$ and $z_t$ that reconstruct these grid maps and sensory signals through a grid-map reconstruction surrogate objective. (b) Two policy learning paradigms with MAD-based agents. Top: Dreamer-style imagination-based policy learning (MAD-Dreamer), where MAD generates rollouts autoregressively and the policy is optimized on imagined trajectories. Bottom: interactive policy learning (MAD-PPO and MAD-SHAC), where MAD acts as a frozen encoder that supplies informative latent representations to a downstream actor network during environment interaction. (c) Deployment of a trained MAD-based policy on a physical quadrotor. The MAD and the actor network are exported as a single policy, deployed on an onboard computer that receives depth images and VIO estimates and outputs control commands to the PX4 flight controller for real-time flight control.}\label{fig:overview}
\end{figure*}

\subsection{Occupancy Grid Map}
Occupancy grid maps are fundamental spatial representations for robot navigation because they explicitly encode which regions are free or occupied. VoxGraph \cite{voxgraph} builds globally consistent volumetric maps with signed-distance submaps, but global alignment and dense updates are costly for high-speed aerial systems. More recent methods reduce this burden: \cite{dmap} avoids expensive ray casting for real-time dense mapping, and ROG-map \cite{rogmap} maintains a robocentric local occupancy map attached to the robot, reducing memory and improving resolution for UAV planning \cite{super,slope}. MAD follows the same robocentric principle, but uses OGM/VGM supervision to shape a learned world model rather than to drive a separate planner.

\section{Methodology}

In this section, we present the methodology of our mapping-aware world model based visual navigation system. The overall framework of the proposed system is depicted in Fig.~\ref{fig:overview}.

We consider a vision-based quadrotor flight task in which a UAV must navigate through cluttered environments using onboard depth sensing and proprioceptive measurements. The GPU-accelerated DiffAero simulator provides depth images, proprioceptive sensory information, and robocentric occupancy and visibility grid maps, which serve as supervision signals for learning a world model. Building on these signals, we design the Mapping-Aware Dreamer (MAD), a latent dynamics model composed of image and sensory encoders, a recurrent state-space model, a sensory decoder, and grid map decoders. MAD is trained in a self-supervised manner to reconstruct local occupancy and visibility grid maps as well as sensory information, thereby learning spatio-temporally consistent latent states that capture both local occupancy and ego-motion. These latent representations are then exploited by two policy learning paradigms: an imagination-based variant, MAD-Dreamer, where MAD generates auto-regressive rollouts for training a policy entirely on imagined latent rollouts, and two interactive variants, namely MAD-PPO and MAD-SHAC, where MAD acts as a frozen encoder that furnishes compact state representations as observations to the actor network during environment interaction. Finally, the trained MAD encoder and actor are exported as a single policy and deployed on an onboard computer, where they process real depth and visual-inertial odometry inputs to output control commands for agile quadrotor flight.

We first formalize the vision-based quadrotor navigation problem and task setup in Section~\ref{sec:problem}. Section~\ref{sec:behaviorlearning} then details the architecture and training of MAD, and Section~\ref{sec:policylearning} describes the policy learning procedures that leverage the learned representations for efficient decision-making and robust control.

\subsection{Problem Statement and Formulation}
\label{sec:problem}
Our objective is to navigate a quadrotor to a specified target position while avoiding obstacles, using only onboard depth sensing, proprioception, and computation. We train policies in the GPU-accelerated DiffAero simulator \cite{zhang2025diffaero} and deploy the learned controller without adding an explicit online mapper or planner.

\subsubsection{Quadrotor Dynamics}
Following  \cite{Zhang2025} and  \cite{zhang2025diffaero}, we use the continuous-time point-mass dynamics model implemented in PyTorch:
\begin{equation}\label{pmc}
\begin{cases}
\dot{\mathbf{p}}=\mathbf v \\
\dot{\mathbf{v}}=\mathbf a+\mathbf g - d\mathbf v \\
\dot{\mathbf{a}}=\lambda(\mathbf u - \mathbf a)
\end{cases}
\end{equation}
where $\mathbf p$ and $\mathbf v$ are the position and velocity in the world frame, $\mathbf g$ is gravity, $\mathbf a$ is the thrust-induced acceleration in the world frame, $d$ is the linear drag coefficient, $\mathbf u$ is the acceleration command, and $\lambda$ approximates actuation and attitude-tracking latency. The commanded attitude is recovered by aligning the body $z$-axis with the desired thrust direction and choosing the yaw from the horizontal velocity direction.

Compared with a full 6-DoF model, this acceleration-limited translational model uses fewer physical parameters while preserving the dominant dynamics relevant to short-horizon collision avoidance \cite{Zhang2025}. It also reduces real-robot calibration effort and, because it is implemented in PyTorch, supports both reinforcement learning and differentiable policy optimization.

\subsubsection{MDP Formulation}
We model vision-based flight as a finite-horizon partially observable Markov decision process (POMDP) $\langle\mathcal O,\mathcal S,\Omega,\mathcal A,F,R,\gamma\rangle$. Here $\mathcal O=\mathcal W\times\mathcal D$ is the observation space, $\mathcal W\subset\mathbb R^{H\times W}$ is the depth-image space, $\mathcal D\subset\mathbb R^{d_\mathrm{sens}}$ is the proprioceptive sensing space, $\mathcal S$ is the environment state space, $\Omega:\mathcal S\rightarrow\mathcal O$ is the observation function, $\mathcal A\subset\mathbb R^{d_\mathrm{act}}$ is the action space, $F:\mathcal S\times\mathcal A\rightarrow\mathcal S$ is the transition model, $R$ is the reward, and $\gamma\in[0,1)$ is the discount factor.

\subsubsection{Observation and Action Spaces}
Observations and actions are expressed in a gravity-aligned local frame whose $x$-axis lies in the vertical plane of the quadrotor body $x$-axis and whose $z$-axis points upward. The observation $o_t\in\mathcal O$ contains a depth image $w_t\in\mathcal W$ with resolution $H\times W=18\times32$ and a proprioceptive vector $d_t\in\mathcal D$. This vector concatenates the clipped target-directed velocity $\mathbf v^\mathrm{tar}\in\mathbb R^3$, the quadrotor velocity $\mathbf v\in\mathbb R^3$, and the local coordinates of the body $z$-axis $\mathbf u_\mathrm z\in\mathbb R^3$, giving $d_\mathrm{sens}=9$. The target-directed velocity is
\begin{equation}
    \mathbf v^\mathrm{tar}=\frac{\mathbf p^\mathrm{tar}-\mathbf p}{\max(||\mathbf p^\mathrm{tar}-\mathbf p||/v_\text{max}, 1)}.
\end{equation}
The action $a_t$ is the desired thrust acceleration in the local frame. It is transformed to the world frame and used as $\mathbf u$ in \eqref{pmc}. During simulation and real flight, this acceleration command and the current velocity are converted to attitude and collective-thrust references for the PX4 low-level controller.

\subsubsection{Reward Function}
As a key component of the RL task, the reward signal should accurately reflect the goal while being dense enough for fast convergence. For the vision-based navigation task specifically, the reward signal is designed as a weighted combination of survival reward $r_\text{srvl}$, collision reward $r_\text{coll}$, velocity reward $r_\text{vel}$, position reward $r_\text{pos}$, height reward $r_\text{z}$, approaching reward $r_\text{app}$, avoiding reward $r_\text{avoid}$, and jerk reward $r_\text{jerk}$ with corresponding weights $\alpha_\text{coll}=5$, $\alpha_\text{vel}=0.06$, $\alpha_\text{pos}=0.5$, $\alpha_\text{z}=0.2$, $\alpha_\text{app}=0.6$, $\alpha_\text{avoid}=0.3$, and $\alpha_\text{jerk}=0.005$:
\begin{equation}
\begin{aligned}
    r&=r_\text{srvl}+\alpha_\text{coll}\cdot r_\text{coll}\\&+\alpha_\text{vel}\cdot r_\text{vel}+\alpha_\text{pos}\cdot r_\text{pos}+\alpha_\text{z}\cdot r_\text{z}\\
    &+\alpha_\text{app}\cdot r_\text{app}+\alpha_\text{avoid}\cdot r_\text{avoid}+\alpha_\text{jerk}\cdot r_\text{jerk}\\
\end{aligned}
\end{equation}
where
\begin{subequations}
\begin{align}
    r_\text{srvl}&=0.3\\
    r_\text{coll}&=-\mathbb I(\text{collision})\\
    r_\text{jerk}&=-||\mathbf u-\mathbf a||_2^2\\
    r_\text{vel}&=-\mathrm{SmoothL1}(||\mathbf v^\mathrm{tar}-\mathbf v||_2, 0)\\
    r_\text{z}&=\mathrm e^{-|p_\mathrm z-p^\mathrm{tar}_\mathrm z|}\\
    r_\text{pos}&=\mathrm e^{-||\mathbf p-\mathbf p^\mathrm{tar}||_2}.
\end{align}
\end{subequations}
For the $i$-th obstacle, we divide the quadrotor's velocity $\mathbf v$ into two parts, $\mathbf v^\mathrm{app}_i$ and $\mathbf v^\mathrm{avoid}_i$, with $\mathbf v^\mathrm{app}_i\parallel \mathbf p^\mathrm{rel}_i$ and $\mathbf v^\mathrm{avoid}_i\perp\mathbf p^\mathrm{rel}_i$, where $\mathbf p^\mathrm{rel}_i=\mathbf p^\mathrm{obst}_i-\mathbf p$ is the relative position of the nearest point of the obstacle $\mathbf p^\mathrm{obst}_i$ with respect to the quadrotor. Then we can define the approaching reward $r_{\mathrm{app},i}$ and the avoiding reward $r_{\mathrm{avoid},i}$ for each obstacle:
\begin{align}
    r_{\mathrm{app},i}\!=\!\begin{cases}
       -||\mathbf v^{\mathrm{app}}_i||_2\!\cdot\!\mathrm e^{-||\mathbf p^{\mathrm{rel}}_i||_2}, &\!\text{if   }\mathbf p_{i}^{\mathrm{rel}\top}\mathbf v^{\mathrm{app}}_i>0\\
        0,&\!\text{otherwise}
    \end{cases}
\end{align}
\begin{align}
    r_{\mathrm{avoid},i}=\begin{cases}
       ||\mathbf v^{\mathrm{avoid}}_i||_2\cdot\mathrm e^{-||\mathbf p^{\mathrm{rel}}_i||_2}, &\text{if   }\mathbf p_{i}^{\mathrm{rel}\top}\mathbf v^{\mathrm{app}}_i>0\\
        0,&\text{otherwise}
    \end{cases}
\end{align}
Then $r_{\mathrm{app}}$ and $r_{\mathrm{avoid}}$ are defined to be equal to $r_{\text{app}_k}$ and $r_{\text{avoid}_k}$ with $k=\arg\min_ir_{\mathrm{app},i}$ being the index of the most dangerous obstacle with the lowest $r_{\mathrm{app},i}$ among all:
\begin{equation}
\begin{aligned}
r_{\mathrm{app}}&=r_{\text{app}_k}\\
r_{\mathrm{avoid}}&=r_{\text{avoid}_k}
\end{aligned}
\end{equation}
With this setting, $r_\text{vel}$, $r_\text{app}$, and $r_\text{avoid}$ are dense enough to guide the agent to learn basic flight skills in the early training stage, while $r_\text{pos}$ provides an extra bonus once the agent approaches the target, reinforcing the trajectory.

\subsection{Learning World Model through Grid Map Reconstruction}
\label{sec:behaviorlearning}
MAD is trained to make the latent state geometrically meaningful. We use two robocentric grid representations: an occupancy grid map (OGM), which marks voxels containing obstacles, and a visibility grid map (VGM), which marks voxels that have been observed by the onboard camera. Unlike a depth image, the pair of maps separates scene geometry from camera visibility and preserves short-term spatial memory. We therefore supervise the world model with OGMs and VGMs rather than using raw depth reconstruction as the primary objective.

\begin{figure}[!t]
\centering
\includegraphics[width=0.48\textwidth]{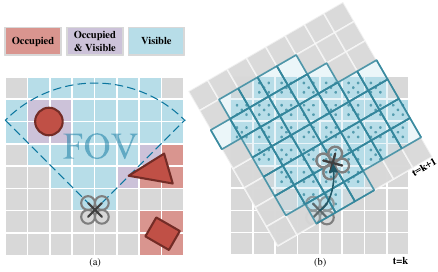}
\caption{\textbf{Illustration of local occupancy and visibility grid maps.} Both maps are defined in the quadrotor-centric local frame. (a) OGMs encode whether each voxel is occupied by obstacles, while VGMs encode whether each voxel has been observed by the on-board depth camera. (b) The visibility re-marking mechanism. Voxels at time step $t=k+1$ will be marked as visible, if they contain at least one anchor point of the visible voxels at time step $t=k$.}\label{occ&vis}
\end{figure}

\subsubsection{Occupancy Grid Maps and Visibility Grid Maps}
OGMs and VGMs are three-dimensional Boolean arrays defined in the robocentric local frame. The OGM $g_t^\mathrm{occ}$ represents obstacle occupancy, while the VGM $g_t^\mathrm{vis}$ represents observation history: a voxel with $g_t^\mathrm{vis}=0$ is unknown, not necessarily free. This distinction is important for safety because the model should not be rewarded for hallucinating occupancy in regions that have never been observed. Aligning the grid with the quadrotor local frame also encourages the latent state to learn heading-normalized spatial structure rather than memorizing global orientations.

We construct OGMs by querying the obstacle geometry at all voxel centers in parallel. In the training environments, obstacles are primitive shapes such as cubes and spheres, which makes the point-in-primitive test highly efficient on the GPU. VGMs combine the current camera frustum with accumulated visibility history, as illustrated in Fig.~\ref{occ&vis}. First, depth pixels are back-projected into rays using camera intrinsics and extrinsics, and voxels intersected by valid rays are marked visible. Second, previously visible voxels are transformed into the current local frame. To reduce aliasing caused by discretization and ego-motion, each previous voxel center is expanded to eight neighboring anchor points, and a current voxel is marked visible if any anchor point falls inside it.

Because the computational complexity grows cubically as the voxel size decreases, we adopt a relatively coarse voxelization compared with grid maps typically used in classical mapping pipelines~ \cite{rogmap}. The local grid map spans $[-3, 5]$, $[-4, 4]$, and $[-2, 2]$ meters along the $x$, $y$, and $z$ axes of the local frame, respectively, with a voxel edge length of $l = 0.4~\mathrm{m}$, resulting in a total of 4,000 voxels. Thanks to the fully GPU-based implementation, the interactive learning environment and grid map computation together can generate up to $1.21 \times 10^5$ environment interactions per second on our workstation, corresponding to approximately $4.84 \times 10^8$ voxel occupancy and visibility evaluations per second. This high throughput enables efficient interaction with the simulator and allows MAD to be trained with massive amounts of data within only a few hours.

\subsubsection{World Model Learning}
MAD follows the recurrent state-space model (RSSM) design of DreamerV3 \cite{hafner2025dreamerv3}, but replaces the main observation-reconstruction target with mapping-aware supervision. It consists of the following components:
\begin{itemize}
    \item \textbf{Recurrent State-Space Model (RSSM)}: The RSSM predicts the recurrent state $h_t$ and the categorical latent state $z_t$. After action $a_{t-1}$, the sequence model updates $h_t$, the prior predicts $\hat z_t$ from $h_t$, and the posterior infers $z_t$ from the encoded observation $x_t=\mathrm{concat}(x_t^\mathrm{img},x_t^\mathrm{snsr})$ when the observation is available.
\end{itemize}
\begin{subequations}
\begin{align}
\quad &\text{Sequence model:} &&\!h_t=\!f_\psi(h_{t-1},z_{t-1},a_{t-1})\\
\quad &\text{Categorical posterior:} &&z_{t}\sim q_\psi(z_t\mid h_t,x_t)\\
\quad &\text{Categorical prior:} &&\hat{z}_t\sim p_\psi(\hat{z}_t\mid h_t)
\end{align}
\end{subequations}
\begin{itemize}
    \item \textbf{Encoders}: Encoder networks map the raw observation $o_t=(w_t,d_t)$ into feature $x_t$ that will be subsequently fed into the categorical posterior $q_\psi$ to evolve categorical latent $z_t$.
\end{itemize}
\begin{subequations}
\begin{align}
&\text{Image encoder:} &x_t^\mathrm{img}&=q_\psi^\mathrm{img}(w_t)\\
&\text{Sensory encoder:} &x_t^\mathrm{snsr}&=q_\psi^\mathrm{snsr}(d_t)
\end{align}
\end{subequations}
\begin{itemize}
    \item \textbf{Decoders}: Unlike the DreamerV3, the decoder networks of MAD reconstruct grid maps $\hat g_t^\mathrm{occ}, \hat g_t^\mathrm{vis}$ and sensory information $\hat d_t$ instead of the exact raw observations $\hat o_t$. Additionally, reward signals and continuation flags are also reconstructed to support the data imagination for policy learning.
\end{itemize}
\begin{subequations}
\begin{align}
&\text{Occupancy decoder:} &\hat g_t^\mathrm{occ}&\sim p_\psi^\mathrm{occ}(h_t,z_t)\\
&\text{Visibility decoder:} &\hat g_t^\mathrm{vis}&\sim p_\psi^\mathrm{vis}(h_t,z_t)\\
&\text{Sensory decoder:} &\hat d_t&\sim p_\psi^\mathrm{snsr}(\hat d_t|h_t,z_t)\\
&\text{Reward decoder:} &\hat r_t&\sim p_\psi^\mathrm{rew}(\hat r_t|h_t,z_t)\\
&\text{Continue decoder:} &\hat c_t&\sim p_\psi^\mathrm{term}(\hat c_t|h_t,z_t)
\end{align}
\end{subequations}
All networks are implemented as multi-layer perceptrons (MLPs), except the image encoder, which is a lightweight convolutional neural network (CNN). Given sequences of observations $o_{1:T}$, actions $a_{1:T}$, rewards $r_{1:T}$, and continuation flags $c_{1:T}$, MAD is optimized using reconstruction, dynamics, and representation losses with weights $\beta_\mathrm{recon}=1$, $\beta_\mathrm{dyn}=0.5$, and $\beta_\mathrm{rep}=0.1$:
\begin{equation}
\begin{aligned}
    \mathcal L(\psi)&\doteq\mathbb E_{q_\psi}\Big[\sum^T_{t=1}\beta_\mathrm{recon}\mathcal L_\mathrm{recon}(\psi)\\
    &+\beta_\mathrm{dyn}\mathcal L_\mathrm{dyn}(\psi)+\beta_\mathrm{rep}\mathcal L_\mathrm{rep}(\psi)\Big]
\end{aligned}
\end{equation}
The reconstruction loss contains grid-map, sensory, reward, and continuation terms:
\begin{equation}
\begin{aligned}
\mathcal L_\mathrm{recon}&\doteq \mathrm{BCE}(\hat g_t^\mathrm{occ}\wedge g_t^\mathrm{vis},g_t^\mathrm{occ}\wedge g_t^\mathrm{vis})\\
&+\mathrm{BCE}(\hat g_t^\mathrm{vis},g_t^\mathrm{vis})-\ln p_\psi(d_t|z_t,h_t)\\
&-\ln p_\psi(r_t|z_t,h_t)-\ln p_\psi(c_t|z_t,h_t).
\end{aligned}
\end{equation}
The first term supervises occupancy only inside visible voxels, because unobserved cells should not be treated as either free or occupied. The KL terms follow the Dreamer-style latent balancing:
\begin{align}
\mathcal L_\mathrm{dyn} &= D_\mathrm{KL}\!\left[\mathrm{sg}(q_\psi(z_t|h_t,x_t))\|p_\psi(\hat z_t|h_t)\right],\\
\mathcal L_\mathrm{rep} &= D_\mathrm{KL}\!\left[q_\psi(z_t|h_t,x_t)\|\mathrm{sg}(p_\psi(\hat z_t|h_t))\right],
\end{align}
where $\mathrm{sg}(\cdot)$ denotes stop-gradient. The OGM/VGM objective is a causal surrogate for depth prediction: it preserves the geometry and visibility needed for flight, while avoiding the need to reproduce raw pixels that are less directly related to collision avoidance.

\begin{figure}[!t]
\centering
\includegraphics[width=0.48\textwidth]{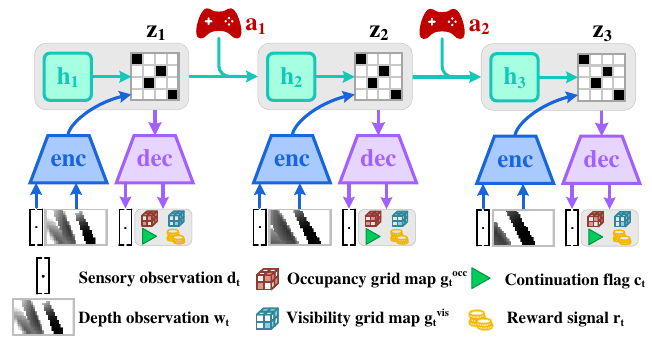}
\caption{\textbf{Training pipeline of MAD.} At each step, MAD encodes the depth observation and sensory inputs into a discrete latent representation $z_t$. The encoder, RSSM state predictor, and decoders are jointly trained to reconstruct the designated targets from $h_t$ and $z_t$, including the sensory observations, occupancy and visibility grid maps, continuation flags, and reward signals.}\label{wm_training}
\end{figure}

The training process is shown in Fig.~\ref{wm_training}. For policy-learning algorithms that do not use imagined rollouts, such as PPO and SHAC, we adopt a task-agnostic recipe: the reward decoder $p_\psi^\mathrm{rew}(\cdot)$ is disabled, and the target-directed velocity $\mathbf v^\mathrm{tar}$ is provided directly to the policy rather than to MAD. Section~\ref{sec:training_results} shows that this encourages transferable visual representations for unseen tasks.

\subsection{Policy Learning}
\label{sec:policylearning}
MAD can be coupled with different policy-learning algorithms without changing the world model  architecture. We evaluate three instantiations: MAD-Dreamer, which uses imagined latent rollouts; MAD-PPO; and MAD-SHAC \cite{schulman2017proximal,xu2021accelerated}. In the feature-extractor setting used by MAD-PPO and MAD-SHAC, the learned MAD encoder is frozen and supplies $h_t$ and $z_t$ as compact observations to the actor.

For imagination-based MAD-Dreamer, we initialize imagined rollouts from short context snippets sampled from a cache of collected trajectories. Unlike earlier generations of Dreamer  \cite{hafner2025dreamerv3}, each rollout starts from a distinct context to ensure broad state coverage and mitigate overfitting to specific motion patterns \cite{hafner2025trainingagentsinsidescalable}.

In our implementation, the context length is set to $4$ and the imagination horizon to 16, offering a balanced compromise between computational efficiency and the capacity to capture long-term dependencies in policy learning. During imagination, MAD predicts latent transitions, rewards, and continuation flags, conditioned on the current latent state and the action. The actor-critic pair is then updated using these imagined trajectories, without requiring further simulator interactions. This process enables efficient exploration within the latent space while maintaining coherent temporal dynamics modeled by MAD.

The actor and critic are defined as :
\begin{subequations}
\begin{align}
\text{Actor:}&& \quad a_t &\sim\pi_{\theta}\left(h_t,z_t\right) \\
\text{Critic:}&& \quad 
v_t&= V_{\theta}(h_t, z_t) \approx 
\mathbb{E}_{\pi_{\theta}, p_{\psi}} \!\left[ 
  \sum_{k=0}^{\infty} \gamma^k r_{t+k} 
\right]
\end{align}
\end{subequations}
where the actor $\pi_\theta$ is implemented as a parameterized Gaussian policy that outputs the mean and variance of the action distribution conditioned on the latent state $z_t$ and the recurrent state $h_t$. The critic $V_\theta$ estimates the expected long-term returns of the imagined trajectories generated under the joint dynamics of the policy $\pi_\theta$ and the world model $p_\psi$, where a discount factor $\gamma=0.997$ is applied. Following DreamerV3, the critic predicts a categorical distribution over bootstrapped $\lambda$-returns, which improves training stability and sample efficiency.  The actor is optimized using a REINFORCE-style gradient estimator with normalized returns. Additional details on the actor–critic objectives and training procedure can be found in  \cite{hafner2025dreamerv3}.

Beyond Dreamer-style imagination training, MAD can therefore serve as a reusable visual state estimator for downstream reinforcement-learning algorithms. This design keeps the policy input compact, improves temporal consistency compared with frame-wise CNN features, and allows the same mapping-aware representation to be reused across navigation and racing tasks. Further details and results are presented in Section~\ref{sec:training_results}.

\section{Experiments}
We evaluate MAD at three levels: training efficiency in DiffAero, closed-loop simulation in Gazebo/PX4, and real-world flight on a physical quadrotor. All training and simulation experiments are performed on a workstation with an Intel Ultra 9-285K CPU and an NVIDIA GeForce RTX 5090 GPU. After training, MAD and the policy network are exported to ONNX and executed as a single real-time inference module.

\subsection{Training Results}
\label{sec:training_results}

\subsubsection{Training Environment}
Training is conducted in DiffAero, an open-source simulator that provides efficient depth rendering and differentiable quadrotor dynamics. We augment DiffAero with OGM/VGM generation to provide self-supervised targets for MAD. The fully PyTorch-based GPU pipeline enables high-throughput interaction and allows each end-to-end training run to finish within a few hours.

\subsubsection{Baseline Comparison}

To evaluate the mapping-aware representation, we compare several policy-learning baselines with their MAD-based counterparts on the visual navigation task. For stable training, the PPO and SHAC critics estimate state values from the full simulator state, while the actors use the designated observations. MAD-PPO and MAD-SHAC replace the depth-image input to the actor with the latent state $z_t$ and recurrent state $h_t$ produced by MAD.

\begin{figure*}[!t]
    \centering
\includegraphics[width=0.85\textwidth]{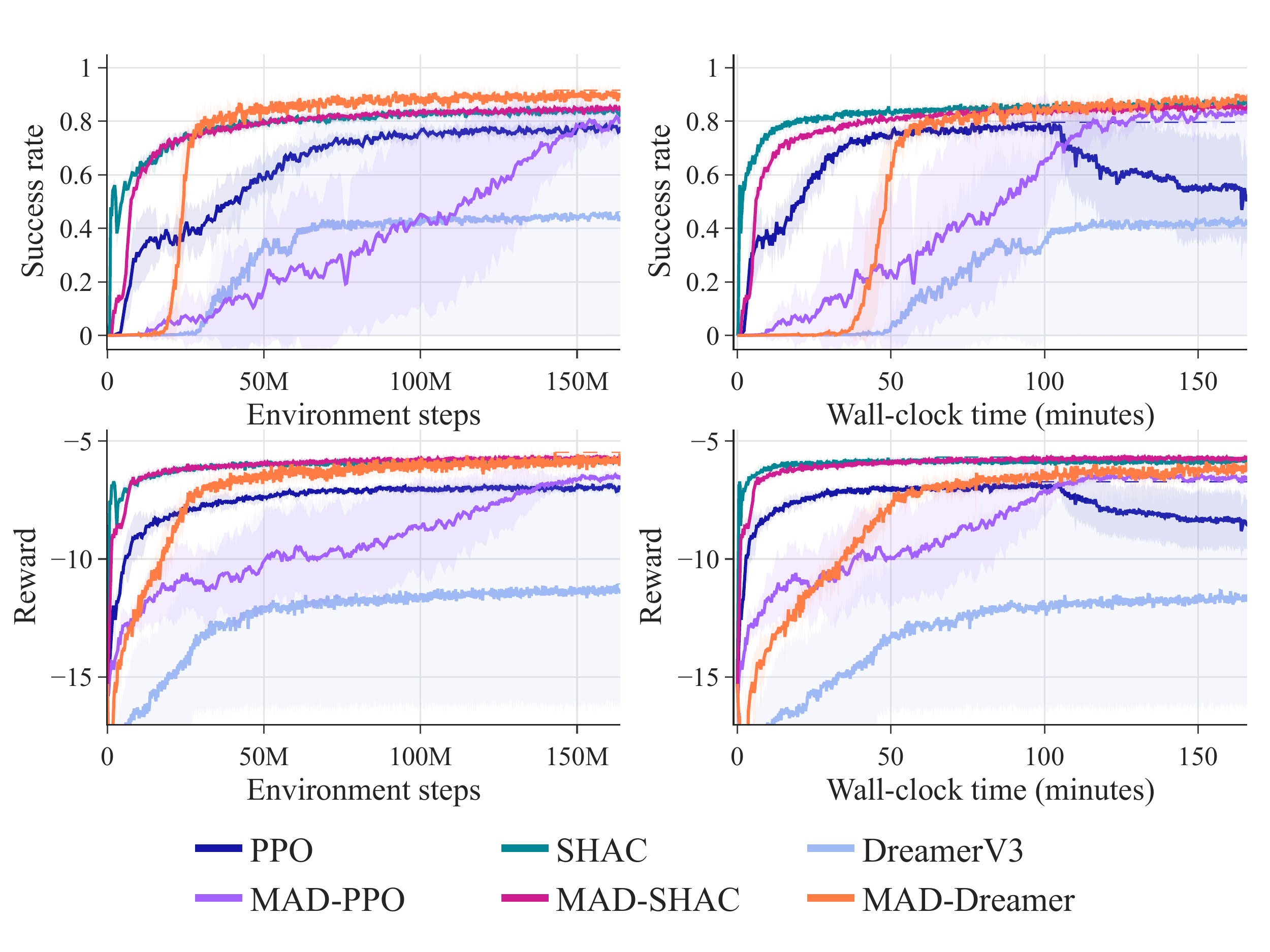}
\caption{\textbf{Training curves of baseline algorithms and their MAD-based variants in the visual navigation task.} Each curve shows the mean episode return and success rates over $8$ random seeds, and the shaded regions indicate the standard deviations across runs.}\label{fig:training_curves}
\end{figure*}

In addition, for MAD-PPO and MAD-SHAC, since the policy networks do not rely on trajectories imagined by MAD, we disable the reward decoder and redirect the target velocity from MAD's encoder to the policy network, as described in Section~\ref{sec:behaviorlearning}, so that MAD focuses on learning task-agnostic representations without concerning itself with predicting task-specific reward signals. Each algorithm is trained with $8$ random seeds in the same training environment, and the resulting learning curves are shown in Fig.~\ref{fig:training_curves}.

As shown in Fig.~\ref{fig:training_curves}, the MAD-based algorithms converge to higher success rates and returns than their non-MAD counterparts, although they require more environment interactions. This warm-up period is expected: reconstructing occupancy and visibility over voxels requires the latent state to learn a richer spatial memory than a policy trained only for immediate control. Because DiffAero generates interactions efficiently, we use a replay ratio of $4$ to reduce overfitting to short-term motion patterns during this representation-learning phase.

\subsubsection{Quality of Reconstructions}
\begin{figure}[!t]
\centering
\includegraphics[width=0.48\textwidth]{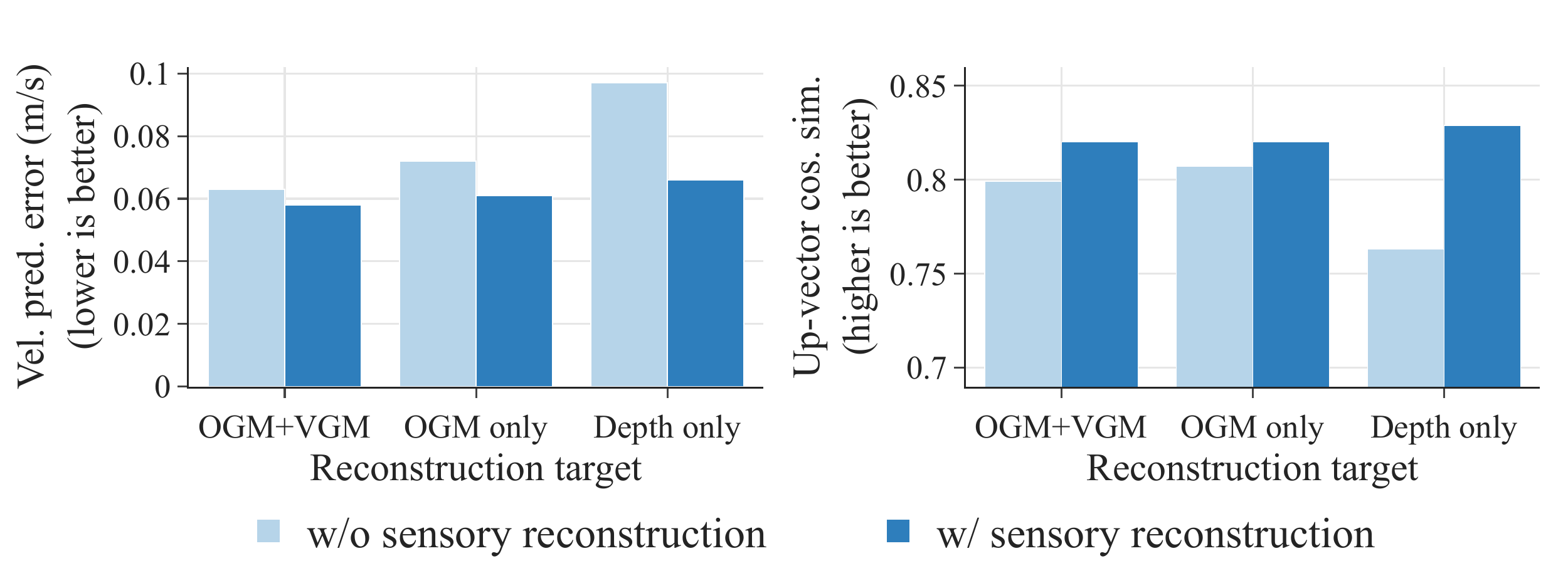}
\caption{\textbf{Velocity and attitude reconstruction errors under different reconstruction strategies.} Darker bars correspond to supervising MAD's sensory decoder to estimate raw sensory inputs using non-detached $h_t$ and $z_t$, allowing gradients to propagate through the entire model. Lighter bars use detached $h_t$ and $z_t$, restricting gradients to the sensory decoder and preventing them from influencing the rest of MAD. The reported result was averaged over $8$ random seeds.}
\label{fig:sensory_reconstruction_error}
\end{figure}

To demonstrate that our grid map reconstruction objective encourages the world model to extract odometry and attitude information from visual inputs, we disable the sensory encoder $p_\psi^\mathrm{snsr}(\cdot)$ so that the only available input is the depth image, and compare the quality of velocity and attitude estimation under different training strategies. We measure the norm of the difference between the estimated and ground-truth local velocity vectors, as well as the cosine similarity between the estimated and ground-truth coordinates of the up-vector in the local frame. The results are depicted in Fig.~\ref{fig:sensory_reconstruction_error}, where each bar aggregates experiments over 8 random seeds.

As shown in Fig.~\ref{fig:sensory_reconstruction_error}, the velocity estimation error is reduced when the reconstruction target switched from depth images to occupancy and visibility grid maps (OGM and VGM), and the accuracy of attitude estimation is also improved. The improvement is especially notable when MAD is not explicitly supervised to produce representations that encode precise velocity and attitude information (bars with lighter blue). These results suggest that, by explicitly training MAD to identify and memorize the currently and previously occupied and visible voxels around the agent, the surrogate grid map reconstruction task encourages MAD to infer the quadrotor's velocity and pose, making it more self-aware of its dynamical state than model-free agents and vanilla world models.
\begin{figure}[!b]
\centering
\includegraphics[width=0.48\textwidth]{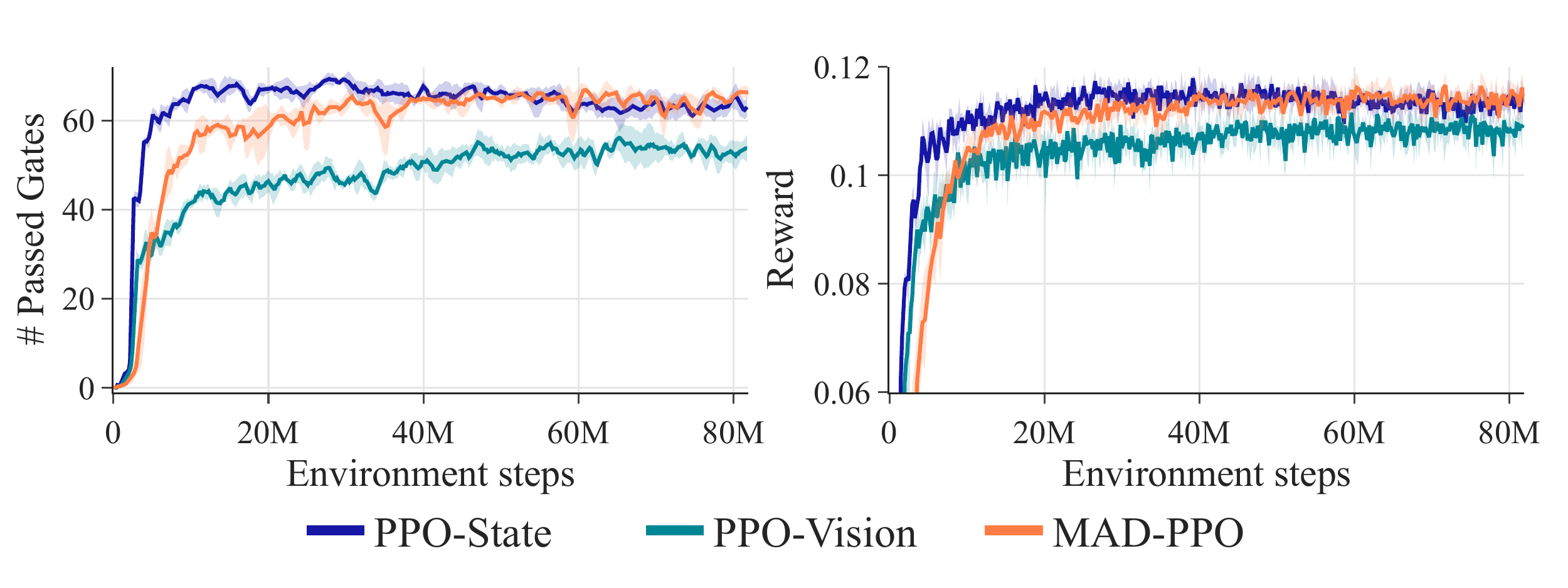}
\caption{\textbf{Learning curves in the vision-based racing task.} MAD-PPO uses a pretrained and frozen MAD as a visual feature extractor and shares the same observation space as PPO-Vision, which is trained from scratch with a CNN-based policy. PPO-State observes only compact state information (positions and poses of upcoming gates) without visual inputs. Each curve is averaged over 4 random seeds.}
\label{fig:racing_training_curve}
\end{figure}

\begin{figure}[!t]
    \centering
    \subfloat[]{\includegraphics[width=0.37\linewidth]{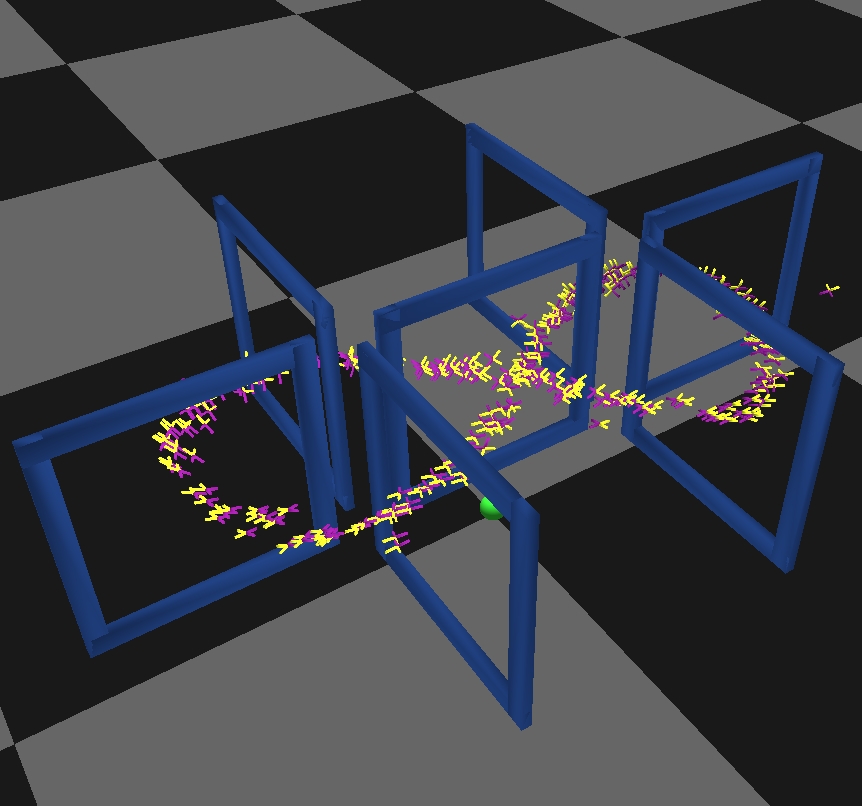}\label{fig:diffaero_racing}}
    \hfill
    \subfloat[]{\includegraphics[width=0.61\linewidth]{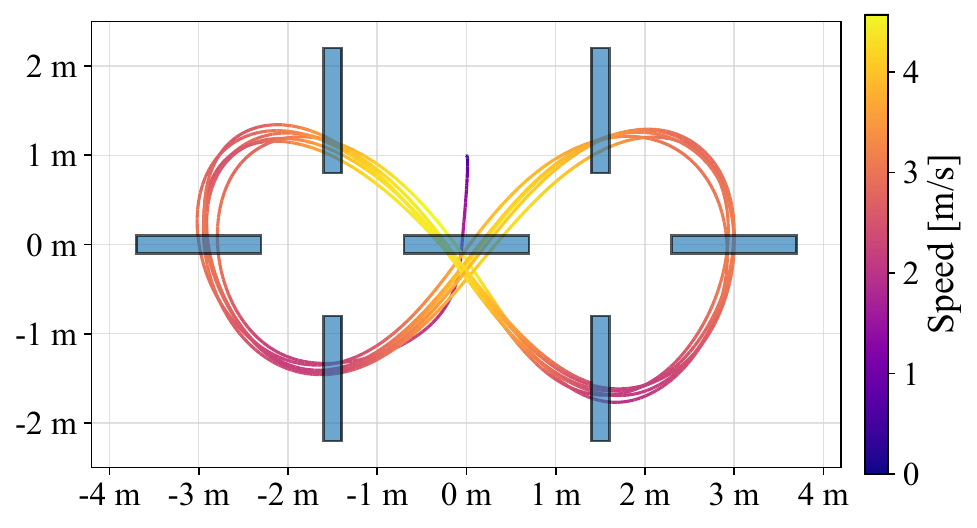}\label{fig:racing_traj_2d}}
\caption{\textbf{Visualization of task transfer to the racing environment.} (a) The vision-based racing task in DiffAero. (b) Top-down view of a flight trajectory controlled by a MAD-PPO agent with a transferred and frozen MAD, demonstrating smooth and agile maneuvers through narrow gates.}
\label{fig:racing_traj}
\end{figure}

\begin{figure*}[!t]
    \centering
        \includegraphics[width=1.1\textwidth]{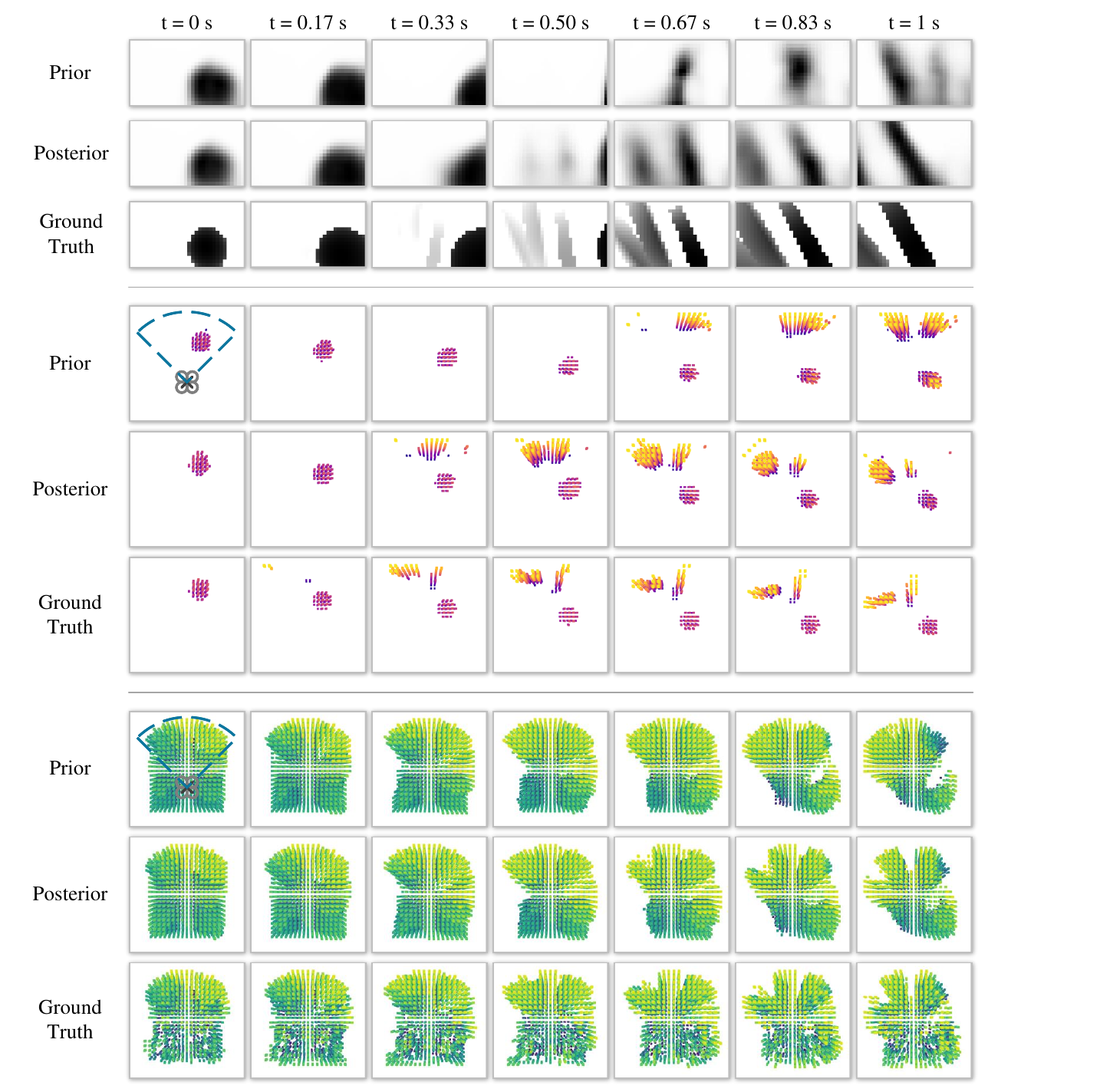}
    \caption{\textbf{Visualization of rollouts imagined by MAD.} Top: depth images predicted by the prior, reconstructed by the posterior, and ground truth from the simulator. Middle: occupancy grid maps predicted by the prior, reconstructed by the posterior, and ground truth. Bottom: visibility grid maps predicted by the prior, reconstructed by the posterior, and ground truth.}
    \label{fig:imagine_rollouts}
\end{figure*}

To further assess the quality of the representations learned by MAD, we visualize depth images, OGMs, and VGMs predicted by the categorical prior, reconstructed by the categorical posterior, and the ground-truth outputs from the simulator, as shown in Fig.~\ref{fig:imagine_rollouts}. The image decoder is trained using detached versions of $h_t$ and $z_t$, preventing gradients from the depth-image reconstruction loss from contributing to the learning of MAD itself. As illustrated in Fig.~\ref{fig:imagine_rollouts}, although MAD is never explicitly trained to produce representations that support depth-image reconstruction, the predicted depth images remain non-trivial and informative, indicating that the learned representations retain sufficient information to reconstruct the visual observations. These observations suggest that the grid map reconstruction task serves as a more effective surrogate than the raw-observation reconstruction objective used in vanilla DreamerV3.

\subsubsection{Task-agnostic Representation Learning}
We transfer a MAD trained on the visual navigation task to a vision-based racing task in order to evaluate the task generalization capability of MAD and its learned representations. We first train a MAD-SHAC agent using the task-agnostic training recipe described in Section~\ref{sec:behaviorlearning}. After convergence, we discard the SHAC policy network and instantiate a new MAD-PPO agent, initializing MAD with the pretrained parameters and randomly initializing the policy network.

Instead of fine-tuning the entire MAD-PPO agent, we freeze the parameters of MAD and use it purely as a feature extractor, feeding $h_t$ and $z_t$ as observations into the PPO policy. We then compare the learning curves of this transferred MAD-PPO agent against two PPO baselines: \emph{PPO-State}, which observes the ground-truth positions and poses of the upcoming gates (without visual inputs), and \emph{PPO-Vision}, which is a vision-based agent trained from scratch with a CNN policy that shares the same observation space as MAD-PPO. Each configuration is trained with 4 random seeds, and the results are reported in Fig.~\ref{fig:racing_training_curve}.

As shown in Fig.~\ref{fig:racing_training_curve}, MAD-PPO with transferred MAD parameters consistently outperforms the PPO-Vision baseline trained from scratch, and gradually even surpasses the state-based PPO-State agent as training progresses. Although MAD is trained only on a single navigation task, these results indicate that its visual feature extraction capability transfers well to structurally different tasks such as racing. The racing environment and a top-down view of the resulting flight trajectory are visualized in Fig.~\ref{fig:racing_traj}, where the transferred MAD-PPO policy performs smooth and agile maneuvers in narrow passages.

\subsection{Simulation Experiments}

\begin{figure*}[!t]
    \centering
    \subfloat[]{\includegraphics[width=0.65\linewidth]{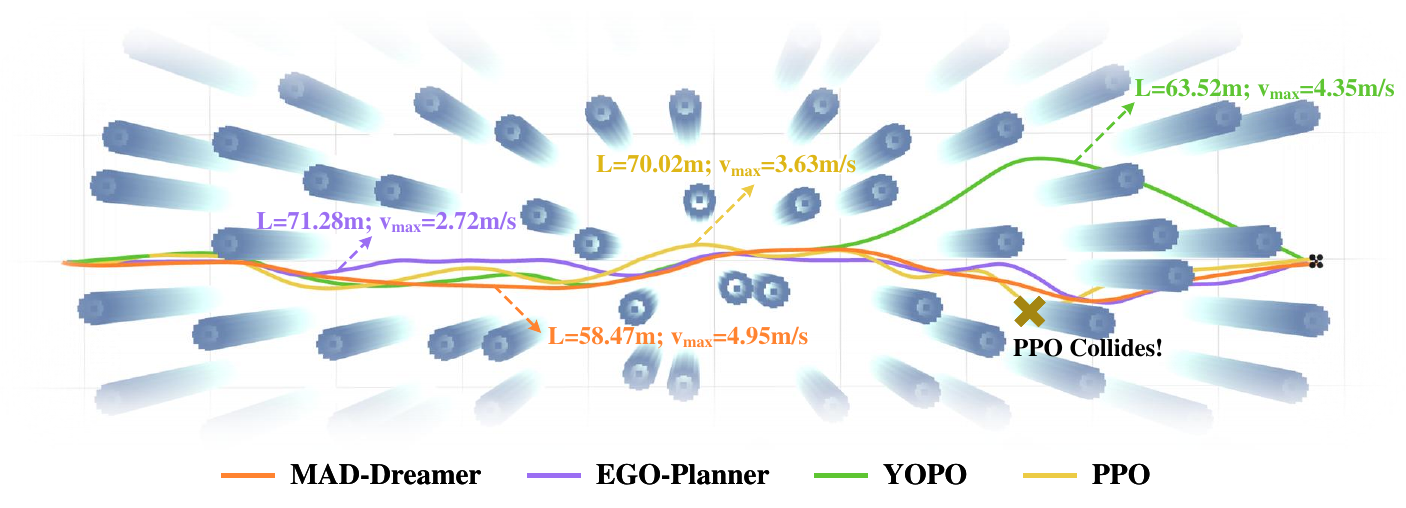}\label{fig:forest_traj}}
    \hfill
    \subfloat[]{\includegraphics[width=0.33\linewidth]{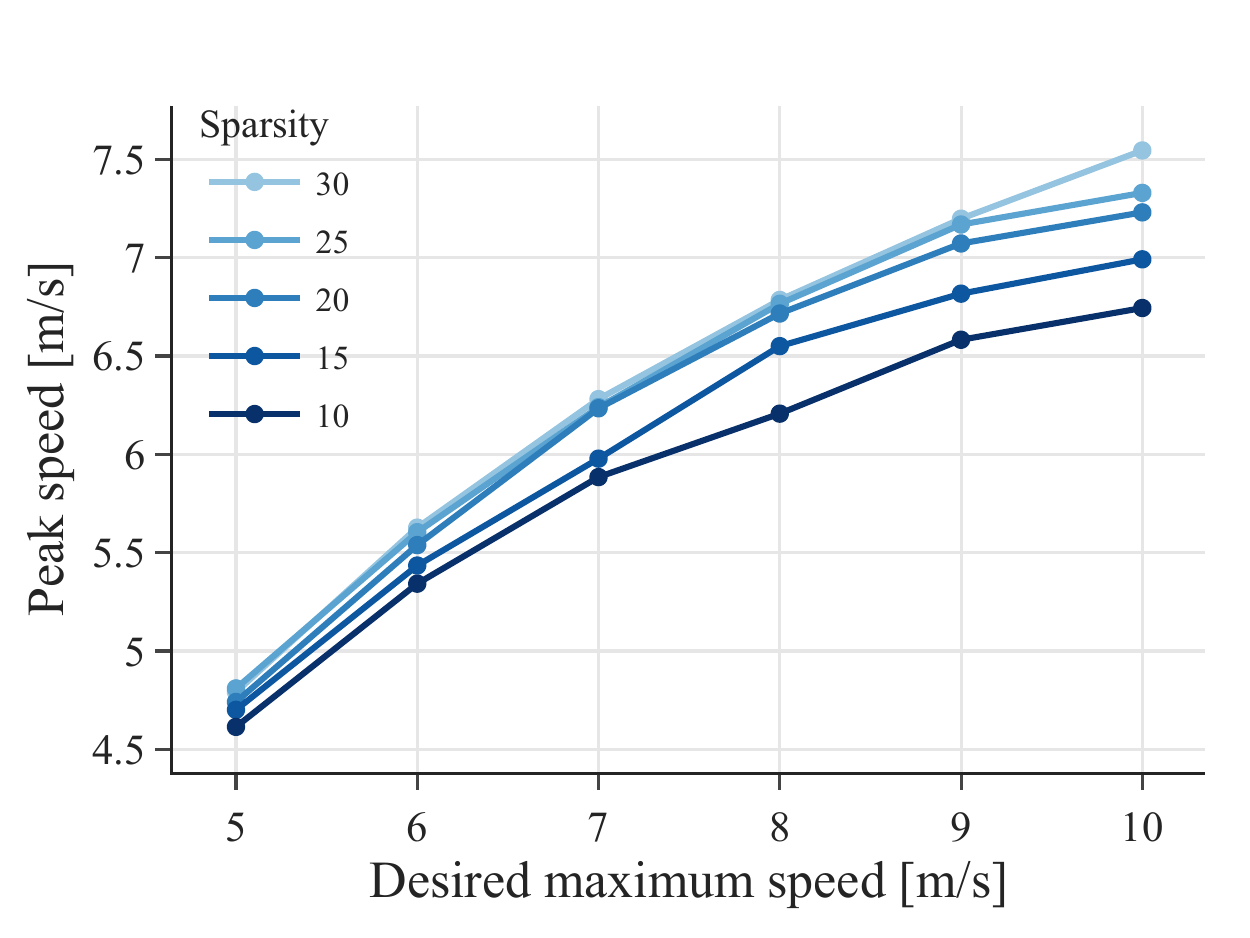}\label{fig:forest_speed_range}}
\vspace{0.3cm}

    \subfloat[]{
    \includegraphics[width=0.242\textwidth]{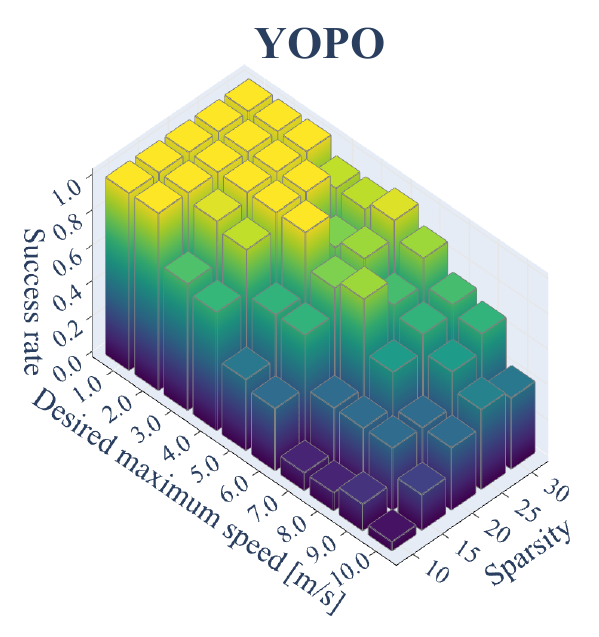}
    \includegraphics[width=0.242\textwidth]{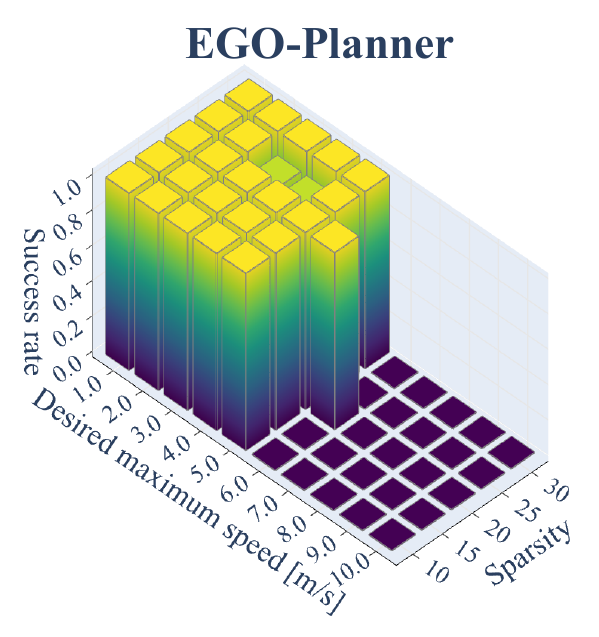}
    \includegraphics[width=0.242\textwidth]{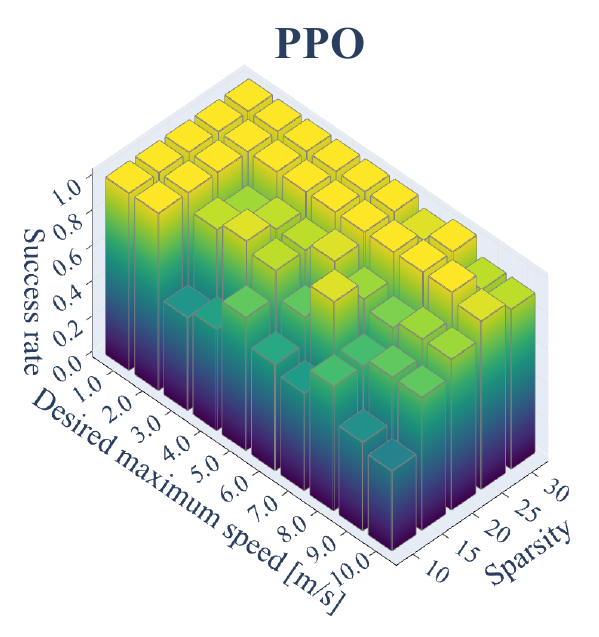}
    \includegraphics[width=0.242\textwidth]{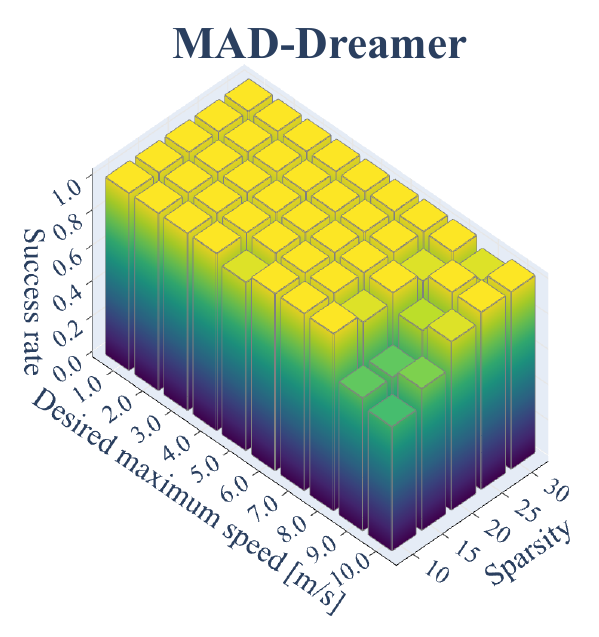}\label{fig:forest_success_rates}
    }
    \caption{\textbf{Simulation performance of baseline algorithms in Gazebo with randomly generated cylindrical obstacles.} (a) Example environment and flight trajectories of YOPO, EGO-Planner, PPO, and MAD-Dreamer for a representative test case. (b) Achieved peak flight speed as a function of the desired maximum speed ($5$-$10$ m/s) and environment sparsity ($10$-$30$ $\text{m}^2/\text{corridor}$). (c) Success rates of YOPO, EGO-Planner, PPO, and MAD-Dreamer across different combinations of environment sparsity and desired maximum speed.}
\label{fig:forest_results}
\end{figure*}

In this section, we present the deployment results of our trained policies in the Gazebo simulator. The experiments are conducted using the PX4 Software-in-the-Loop (SITL) framework, which provides a realistic flight-control pipeline, consistent MAVLink interfaces, and accurate actuator and sensor emulation, thereby reducing the simulation-to-real discrepancy. We evaluate MAD-Dreamer against EGO-Planner \cite{egoplanner}, YOPO \cite{yopo2025}, and PPO in the same simulation environment. All baselines are carefully tuned to achieve balanced real-time performance and safety for fairness. For EGO-Planner specifically, we employ a tracking MPC to follow its generated B-spline trajectories. For fairness, all methods are evaluated under a limited sensing range of approximately $5$ m, consistent with the capabilities of real depth camera hardware.

We first construct a forest-like simulation environment composed of multiple cylindrical obstacles, each with a radius of approximately $0.4$ m. The difficulty level of this environment is controlled by a sparsity parameter, which specifies the area exclusively occupied by one obstacle. For example, a sparsity value of $10$ means that the whole area is partitioned into squares with an area of $10\text{ m}^2$ and generates one obstacle for each square. We consider five difficulty levels with sparsity values of $10$, $15$, $20$, $25$, and $30$, covering environments ranging from dense to relatively sparse obstacle distributions. To ensure statistical robustness, we generate $20$ randomly generated forest scenes for each sparsity value and evaluate different methods at varying desired velocities from $0$ to $10$ m/s across all scenes.

Based on the results shown in Fig.~\ref{fig:forest_success_rates}, the proposed MAD-Dreamer consistently outperforms the baselines across all sparsity levels and across different desired maximum velocities. Its performance remains stable and reliable even in dense environments in which the frequent collision risks and rapid replanning demands make high-speed flight extremely challenging for both classical and learning-based algorithms. In contrast, the performance of EGO-Planner, YOPO, and the PPO-trained policy degrades as the environment becomes denser or as the desired velocity increases.
The performance drop can be attributed to the depth camera's limited sensing range and field of view. Classical model-based planners suffer from incomplete observation of the environment, resulting in local minima and overly conservative trajectories in dense environments. Similarly, YOPO directly regresses trajectories from depth images with a fully convolutional network. Still, it lacks an internal mechanism to aggregate temporal information, causing degraded decision-making in cluttered scenes. In contrast, MAD's latent representation—shaped by its grid map reconstruction task—carries richer spatial and temporal information, leading to improved robustness against widely distributed environment sparsity. This enables reliable navigation with limited visual observations even at high speed. 

Fig. \ref{fig:forest_traj} illustrates representative trajectories generated in a dense forest environment with one cylinder for every $9$ m$^2$ spaces. All algorithms are tested with a desired maximum speed of $5$ m/s. In such cluttered settings, MAD consistently generates smooth and collision-free trajectories, achieving the highest peak speed of $4.95$ m/s and the shortest overall path length of $58.47$ m, while still actively reacting to rapidly emerging obstacles. The flight speed analysis in Fig.~\ref{fig:forest_speed_range} further illustrates how a MAD-based agent adapts its behavior across environments with different sparsity levels. For a fixed desired maximum velocity, MAD-Dreamer produces markedly different peak speeds depending on obstacle density: in dense forests, the policy deliberately lowers its velocity to ensure safety, whereas in sparse environments it fully exploits the agility and attains much higher peak speeds. This trend confirms that our MAD-based policy learned non-trivial risk-aware behavior rather than blindly tracking the desired velocity. Moreover, the rise of peak speed with increasing sparsity reflects MAD's ability to infer local traversability from its learned spatial representation. Since MAD is supervised to extract features that contain local temporal occupancy and visibility status rather than raw depth images, the feature can be exploited by the actor to reason about whether a space is occupied or free and thus perform more reliable actions. This spatially grounded understanding enables both conservative traverse in cluttered regions and agile maneuvers in open areas.

\begin{figure}[!t]
\centering
    \includegraphics[width=1\linewidth]{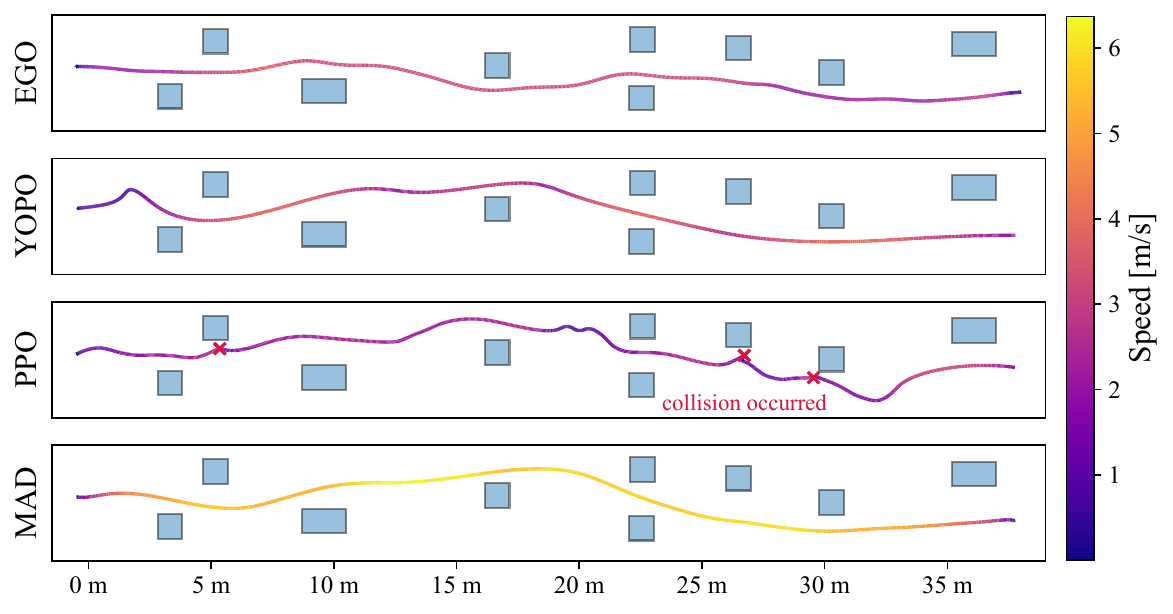}
    \caption{\textbf{Flight trajectories of MAD-Dreamer and three baseline algorithms in an indoor corridor environment.} Under the same sensing range, MAD-Dreamer exhibits markedly smoother and more agile trajectories than all baselines.}
    \label{fig:corridor_traj}
\end{figure}

\begin{figure}[!b]
    \centering
    \includegraphics[width=1.02\linewidth]{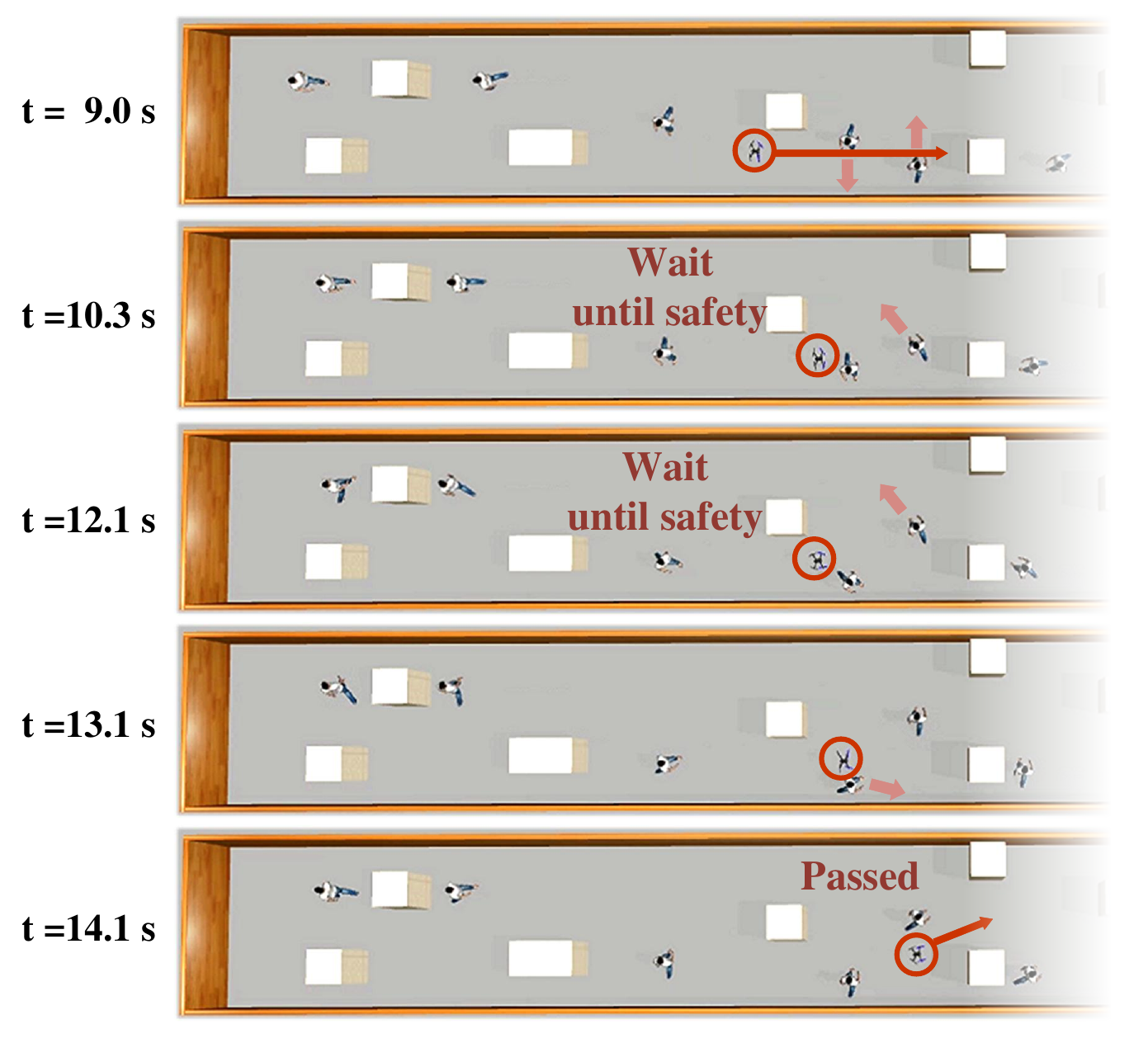}
    \caption{\textbf{Safe navigation visualization in a simulated dynamic environment.} The dynamic scenario demonstrates the MAD-based agent's robust capability for timely and preventive avoidance of moving obstacles (pedestrians).}
    \label{fig:dynamic_obstacle}
\end{figure}

To further evaluate deployment feasibility, we construct an indoor obstacle-avoidance benchmark in Gazebo following the design principles of NavRL \cite{navrl}, consisting of a $5$ m$\times$$40$ m corridor populated with irregularly spaced obstacles. We tune the configurations of YOPO, EGO-Planner, PPO, and MAD-Dreamer to obtain strong and stable performance for each method. As shown in Fig.~\ref{fig:corridor_traj}, MAD-Dreamer achieves the shortest overall completion time of $8.36$ s, significantly outperforming EGO-Planner ($14.33$ s) and YOPO ($16.02$ s). In addition, our agent reaches a peak speed of $6.37$ m/s, whereas EGO-Planner and YOPO attain $4.01$ m/s and $3.71$ m/s, respectively. PPO, in contrast, fails to navigate in this constrained indoor setting and frequently collides, which shows the limit of the purely task-oriented end-to-end policy learning method in cluttered environments where the ability of spatial reasoning is crucial. YOPO's degradation in indoor environments primarily stems from its reliance on global ESDF and point-cloud supervision during training, which induces strong domain dependence on the outdoor-forest setting. In contrast, MAD's grid-map-centric latent representation imposes explicit spatial structure and temporal coherence, enabling more reliable cross‑domain transfer to previously unseen indoor layouts. Our goal is not to assert universal superiority on all benchmarks, but to show that explicitly structured spatio-temporal representations offer greater robustness for zero-shot transfer under substantial domain shifts. 

We further challenge MAD-Dreamer in a dynamic environment with moving obstacles. As shown in Fig.~\ref{fig:dynamic_obstacle}, although the agent is not trained or fine-tuned in dynamic scenes, it reacts to the approaching obstacle by decelerating and stopping before a potential collision. This qualitative result suggests that the spatially structured MAD representation can support reactive safety behavior beyond the static training distribution.

\begin{figure}
    \centering
    \includegraphics[width=1\linewidth]{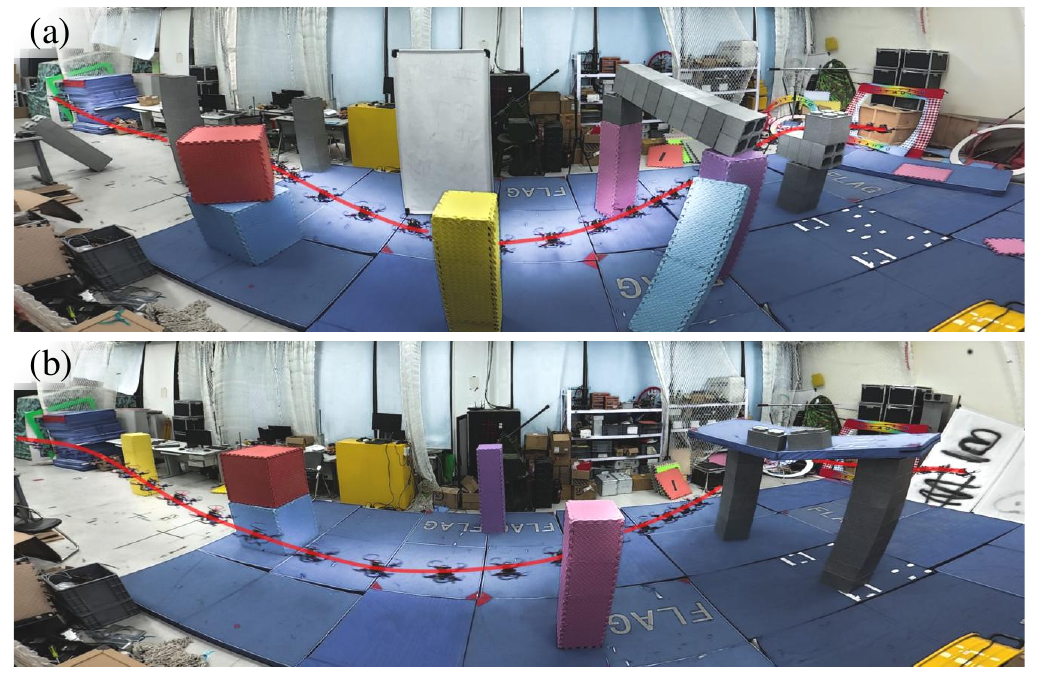}
    \caption{\textbf{Indoor flight trajectories of the proposed MAD.} (a) Scenario I, a more cluttered environment where the quadrotor maintains relatively slow yet safe navigation ($2.06$ m/s).(b) Scenario II, where the obstacles are placed more coarse. Under the control of a MAD-Dreamer agent, the quadrotor flies at a higher speed ($3.10$ m/s).}
    \label{fig:indoor_traj}
\end{figure}

\subsection{Real-world Experiments}

\begin{figure}[!b]
    \centering
    \includegraphics[width=1\linewidth]{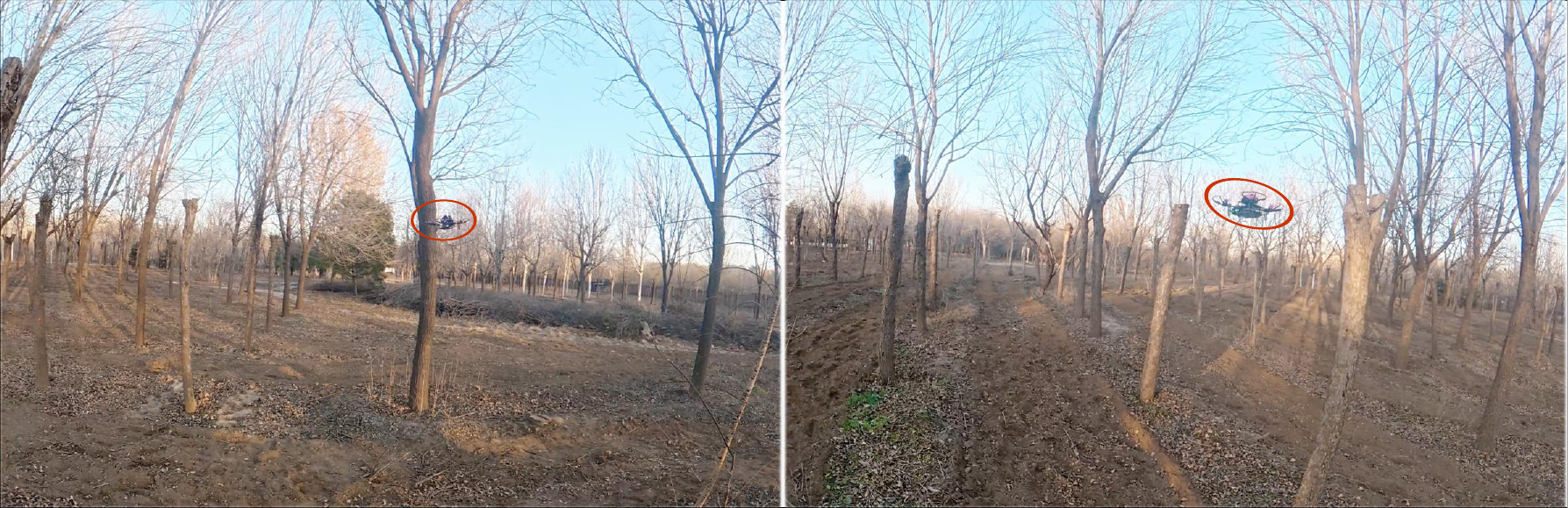}
    \caption{Outdoor flight experiments in a cluttered forest, where the quadrotor maintains safe navigation at speeds up to 5.05 m/s under the control of a MAD-Dreamer agent.}
    \label{fig:outdoor}
\end{figure}

We validate MAD in indoor and outdoor environments on a compact quadrotor with a $250$ mm wheelbase and total mass of $1.10$ kg. The platform uses GEMFAN 5-inch propellers and T-Motor F60 PRO IV 2550 KV motors, giving a thrust-to-weight ratio of $4.2$. It is controlled by an NxtPX4v2 flight controller and a MicoAir 33A 4-in-1 ESC, with an ASUS NUC equipped with an Intel Core i7-1260p CPU as the onboard computer. The exported ONNX model runs with approximately $4$ ms inference latency. Perception is provided by an Intel RealSense D435i depth camera at $30$ Hz with an $\text{87}^\circ\times\text{58}^\circ$ field of view.

To evaluate sim-to-real transfer and assess the safety of our approach, we first performed physical flight experiments in challenging indoor environments. As shown in Fig.~\ref{fig:indoor_traj}, MAD maintains stable trajectories through highly cluttered spaces with irregular obstacle layouts and narrow corridors, achieving safe and consistent flight at speeds up to $3.10$ m/s.

We then tested MAD in a disturbed outdoor setting. As illustrated in Fig.~\ref{fig:outdoor}, the experiments were conducted in a dense forest with a sparsity of roughly $10$, further complicated by persistent wind disturbances that noticeably perturb the quadrotor. Despite these conditions, the quadrotor executes safe trajectories through the cluttered forest and reaches a top speed of $5.05$ m/s.

Together, these indoor and outdoor experiments verify that MAD-based policies transfer from simulation to real flight while maintaining safe navigation behavior under limited onboard sensing.


\section{Conclusions and Discussions}

This paper presented MAD, a mapping-aware world model for agile vision-based quadrotor flight. MAD replaces raw-observation reconstruction with robocentric occupancy and visibility reconstruction, forcing the recurrent latent state to encode local geometry, observation history, and ego-motion. With GPU-parallel OGM/VGM generation in DiffAero, the model can be trained at high throughput and then coupled with Dreamer-style imagination learning, PPO, or SHAC. The experiments show that MAD-based agents improve training performance, provide interpretable map predictions and ego-motion estimates, transfer to a vision-based racing task, and execute safe high-speed flight in Gazebo and on a real quadrotor under limited sensing.

Several limitations remain. The grid-map construction module and the associated reconstruction losses introduce non-trivial memory overhead, which limits the current training scale. In addition, the downstream policy uses the reconstructed representation implicitly and does not yet perform explicit feasibility checking or safety filtering on the decoded maps. This means that the learned controller remains reactive at execution time, even though its internal state is spatially structured.

Overall, the results suggest that mapping-aware world models provide a practical middle ground between classical modular navigation and monolithic end-to-end policies. Future work will investigate tighter integration with trajectory optimization, probabilistic occupancy prediction, and explicit safety filters that can exploit the decoded OGM/VGM information to generate dynamically feasible and certifiably safer flight commands. Beyond depth-based navigation, MAD also points toward world models with richer spatial memory and object-level structure for more general aerial autonomy.




\bibliography{references}
\bibliographystyle{IEEEtran}

\end{document}